\documentclass[letterpaper, 10 pt, journal, twoside]{IEEEtran}
\IEEEoverridecommandlockouts

\usepackage{cite}

\usepackage{url}

\usepackage{amsmath,amssymb,amsfonts}
\usepackage{algorithmic}
\usepackage{graphicx}
\graphicspath{ {./images/} }

\usepackage{tabularx, booktabs}
\usepackage{multirow}
\usepackage[footnotesize]{caption}
\newcommand\boldblue[1]{\textcolor{blue}{\textbf{#1}}}

\usepackage{dirtytalk}

\usepackage{algorithm}
\usepackage{footnote}

\usepackage[caption=false, font=footnotesize]{subfig}

\usepackage{etoolbox}

\usepackage{gensymb}
\usepackage{textcomp}
\usepackage{xcolor}

\def\BibTeX{{\rm B\kern-.05em{\sc i\kern-.025em b}\kern-.08em
    T\kern-.1667em\lower.7ex\hbox{E}\kern-.125emX}}
\begin{document}

\pagenumbering{gobble} 

\title{Marker-Based Localisation System Using an Active PTZ Camera and CNN-Based Ellipse Detection}

\author{Xueyan Oh, Ryan Lim, Shaohui Foong, and U-Xuan Tan

\thanks{This research is supported by the National Robotics R\&D Programme Office, Singapore and Agency for Science Technology and Research (A*STAR), Singapore under its National Robotics Programme - Robotics Enabling Capabilities and Technologies (W1925d0056).}
\thanks{The authors are with the Engineering Product Development Pillar, Singapore University of Technology and Design, Singapore
        {\tt\footnotesize xueyan\_oh@mymail.sutd.edu.sg}}
}

\makeatletter
\patchcmd{\@maketitle}
  {\addvspace{0.5\baselineskip}\egroup}
  {\addvspace{-2\baselineskip}\egroup}
  {}
  {}
\makeatother

\maketitle

\begin{abstract}
Localisation in GPS-denied environments is challenging and many existing solutions have infrastructural and on-site calibration requirements. This paper tackles these challenges by proposing a localisation system that is infrastructure-free and does not require on-site calibration, using a single active PTZ camera to detect, track and localise a circular LED marker. We propose to use a CNN trained using only synthetic images to detect the LED marker as an ellipse and show that our approach is more robust than using traditional ellipse detection without requiring tuning of parameters for feature extraction. We also propose to leverage the predicted elliptical angle as a measure of uncertainty of the CNN's predictions and show how it can be used in a filter to improve marker range estimation and 3D localisation. We evaluate our system's performance through localisation of a UAV in real-world flight experiments and show that it can outperform alternative methods for localisation in GPS-denied environments. We also demonstrate our system's performance in indoor and outdoor environments.
\end{abstract}

\begin{IEEEkeywords}
Localisation, AI and Machine Learning
\end{IEEEkeywords}

\vspace{-3mm}

\section{Introduction}

Localisation is a fundamental capability of autonomous robots but localisation in GPS-denied environments is still challenging \cite{R4}. Common on-board methods that tackle this include the use of LiDAR, optical flow, or fused methods such as Visual-Inertia Odometry (VIO) can be effective in the short term but require a feature-rich environment and can suffer from drift over time \cite{R3}. They also increase the weight and on-board computational requirements which are unfavourable for smaller Unmanned Ground Vehicles (UGVs) or Unmanned Aerial Vehicles (UAVs) \cite{R4}. Off-board methods such as Radio Frequency (RF) and ultra-wideband (UWB) require on-site setup, calibration and can be unreliable due to interference, reflection, or degradation of signals by surrounding obstacles \cite{R6}. Additionally, UWB requires increasing the separation between anchors for localisation at further distances.

Commercial motion capture systems such as Vicon \cite{R7} and OptiTrack \cite{R8} and other multi-camera systems \cite{R9}  can provide very high accuracy but require setup of extensive infrastructure, on-site calibration, and can be very expensive. To address these challenges, monocular approaches \cite{R10,R11} use a single camera to detect and localise fiducial markers such as ArUco \cite{R1} attached onto robots. However, such markers are required to be large and unobstructed for reliable detection at further distances and can affect the aerodynamics of UAVs \cite{R13}. Marker-less monocular detection and localisation of robots have also been explored \cite{R14} but still requires the robot’s size to be known beforehand and often fails in unstructured environments with complex backgrounds.

Additionally, the localisation range of the above off-board visual methods are limited by the scale of their setup and camera’s resolution. The use of Pan-Tilt-Zoom (PTZ) cameras can potentially address these limitations with their ability to zoom and change viewpoint. There are many studies that use PTZ cameras, including for UAV tracking \cite{R16}, people tracking \cite{R18} and surveillance \cite{R19}, but few works explore the use of a single PTZ camera for precise 3D localisation of robots. You et al. \cite{R22} combine PTZ target tracking with cell features on the floor for robot localisation but have only been tested at a small room scale. Unlu et al. \cite{R23} and Oh et al. \cite{R24} use a single PTZ camera and use its zoom capability to extend localisation range. Both methods use traditional background subtraction for moving object detection requiring manual tuning of parameters, before using deep learning to detect a UAV and estimate its size. These methods have also yet to be extensively evaluated in uncontrolled environments.

\begin{figure}[t]
\centerline{\includegraphics[width=1.0\columnwidth]{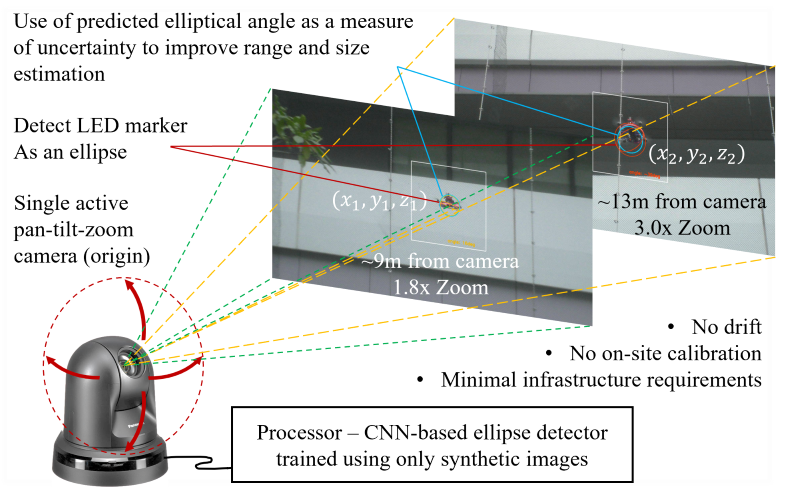}}
\vspace{-1mm}
\caption{Our proposed easily deployable off-board visual localisation system.}
\label{fig:setup}
\vspace{-7mm}
\end{figure}

Accurate and reliable detection of a target in dynamic scenes due to the constantly changing viewpoint of a moving camera is challenging. Existing research has focused on background modelling \cite{R25,R26} across the entire pan-tilt range of a PTZ camera to enable background subtraction at every viewpoint but it is still very challenging to account for variations in lighting conditions and dynamic environments \cite{R23}. Fiducial markers \cite{R1} have also been used for visual localisation without the need for background subtraction but also do not perform well under difficult lighting, cluttered background, and motion blur. To address these challenges, we train a Convolutional Neural Network (CNN) to track a circular LED marker with an active PTZ camera and propose to leverage its predicted elliptical angle as a measure of uncertainty, to filter noisy CNN predictions and improve range estimation. The main contributions of our work are summarised as follows:

\begin{itemize}
\item We propose an easily deployable external localisation system that uses a single ground-based PTZ camera to actively track and localise a circular LED marker. We detect this marker as an ellipse and predict its parameters by training a CNN using synthetic images. We show that our CNN can more robustly detect the LED marker without requiring tuning of parameters for feature extraction as compared to using traditional ellipse detection.

\item We propose to leverage the predicted elliptical angle as a measure of uncertainty in the CNN's predictions and show how it can be used to filter large noises in marker range estimation and improve 3D localisation accuracy.

\item We demonstrate the capability of our proposed system through localisation of a UAV in real-world environments. We evaluate its performance by comparing to existing methods for localisation in GPS-denied environments and show that our method can achieve better localisation accuracy, does not suffer from drift, and does not require any additional infrastructure or on-site calibration.
\end{itemize}

\section{Related Work}

\subsection{Marker-Based Visual Localisation}
Fiducial markers such as ArUco \cite{R1} are often used for visual localisation. Relevant to our work is the attachment of such markers onto mobile platforms and tracking them using external camera(s). For example, Silva et al. \cite{R29} use an upward-facing monocular camera to detect and track an ArUco marker attached to the bottom of a UAV for vision-based landing. Pickem et al. \cite{R30} attach a fiducial marker on top of each ground robot for multi-robot localisation within a testbed using an overhead webcam. Tsoukalas et al. \cite{R31} propose a complex 3D arrangement of fiducial markers attached to the top of UAVs for close-range (2.2 to 5 m) relative pose estimation. However, the use of these markers is limited by constraints of the robotic platform, such as robot size, the position of other payloads, or the aerodynamics of UAVs \cite{R13}.

LED markers are also used for tracking and localisation tasks \cite{R35}. While active LED markers blinking at high frequencies can be robustly detected using cameras with high FPS, they cannot be detected during their off-phase \cite{R36} and cannot be used with many industrial PTZ cameras due to their lower FPS (such as 12.5 FPS on a Panasonic HE40 at HD resolution). Non-blinking LEDs are better suited for visual systems with lower frame rates. For example, Jin and Wang \cite{R37} mount four round LED markers onto a UAV and detect them using a static camera for localisation but have only explored varying distances in one dimension along camera's direction. Sun et al. \cite{R38} use LED arrays for optical camera communication between vehicles and focus on LED segmentation recognition by combining traditional image processing and machine learning. Our work uses an active PTZ camera to track a non-blinking LED marker for 3D localisation.

Elliptical markers have also been used for localisation tasks. For example, Jin et al. \cite{R61} and Keipour et al. \cite{R62} both detect a black-and-white elliptical marker from a UAV to perform autonomous landing. However, they require the black elliptical marker to be against a white background and use traditional image processing methods requiring manual tuning of parameters for feature extraction. Recently, Stuckey et al. \cite{R63} use a single off-board camera to track a circular LED marker attached to a UAV for 3D localisation. However, they also use traditional feature extraction and have only tested in an indoor controlled environment. There are also existing deep learning-based ellipse detectors \cite{R44, R64} but are mostly trained to detect multiple generic elliptical objects within images in non-dynamic conditions. Our approach also removes the need to tune parameters for feature extraction by training a CNN to specifically detect a single circular LED marker as an occluded ellipse (observed mostly as an arc or even a straight line - see Fig.~\ref{fig:trad_vs_ours}) and predict its elliptical parameters. We also propose to leverage the predicted elliptical angle as a measure of uncertainty and use it to filter noisy predictions by the CNN.

\subsection{UAV Localisation Using PTZ Cameras}
Few studies have explored the use of PTZ cameras to localise UAVs. Unlu et al. \cite{R23} compare different background subtraction models and use a k-nearest-neighbour (KNN) model to search for bounding boxes of moving objects within an active PTZ camera’s view and use a CNN for identification. They use the detected width of the UAV to estimate its location and report a root-mean-square error (RMSE) of 0.67m for an indoor experiment. Oh et al. \cite{R24} train a CNN using synthetic binary images to detect and predict the size and position of a custom LED marker (attached to the underside of a UAV) in PTZ camera frames for 3D localisation as well as demonstrate localisation at different manually-set zoom levels. However, this work also uses background subtraction to obtain the binary images as the CNN’s input. These works also emphasise the importance of size estimation for determining ranging distance in monocular 3D localisation. Our approach does not require background subtraction, can work with dynamic background, and reduces the effect of noisy size estimations on distance estimates to improve 3D localisation accuracy.

\begin{figure*}[t]
\centering
\centerline{\includegraphics[width=1.0\textwidth]{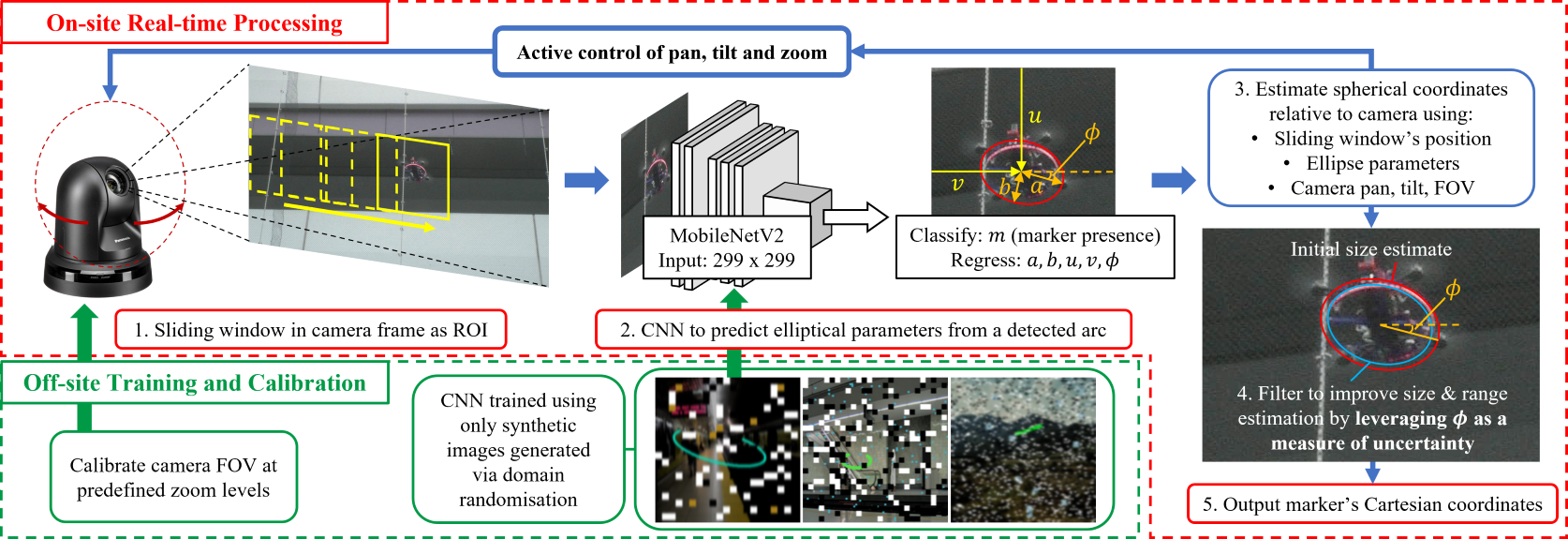}}
\vspace{-1mm}
\caption{Overview of our proposed localisation system. Our system is capable of real-time detection, tracking and localisation of a circular LED marker by 1) extracting a ROI; 2) using the ROI as input into a CNN trained using purely synthetic images for detection and prediction of an ellipse’s parameters; 3) active control of the PTZ camera (with calibrated FOV at predefined zoom levels) for tracking the marker while estimating the marker’s Spherical coordinates relative to the PTZ camera; 4) leveraging the predicted elliptical angle as a measure of uncertainty in the CNN's predictions and using it in an adaptive particle filter to improve range estimates; and 5) update state estimate and convert to Cartesian coordinates as output.}
\label{fig:system}
\vspace{-6mm}
\end{figure*}

\section{Proposed Localisation System}

\subsection{Overview of Localisation System}
We propose an easily deployable off-board localisation system that consists of a single PTZ camera at a fixed position on the ground and a processor to detect, actively track, and localise a circular LED marker. We propose to use a CNN to detect this marker and predict its elliptical parameters, and generate synthetic images to train this CNN. We split the PTZ camera's zoom into multiple pre-defined levels with known field-of-view (FOV). We then use an adaptive particle filter to improve the distance estimate of the marker from the camera. Lastly, we implement PTZ camera control for active tracking of the marker and estimate its Cartesian coordinates. Fig.~\ref{fig:setup} shows the setup of our proposed off-board visual localisation system and Fig.~\ref{fig:system} shows an overview of the system.

\subsection{Circular LED Marker}\label{ledmarker}
\begin{figure}[t]
\centerline{\includegraphics[width=1.0\columnwidth]{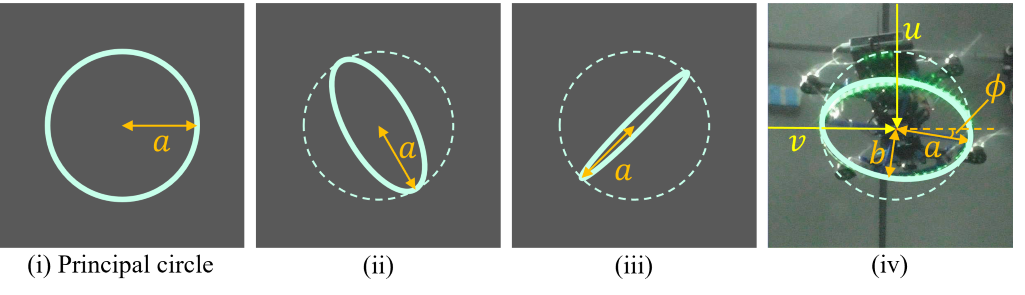}}
\vspace{-1mm}
\caption{We leverage the concept that projections of a circle, at the same position but different orientations relative to an image plane, are ellipses and can all be approximated to have the same semi-major axis length $a$ and centroid position. In this example, ellipses (ii), (iii) and (iv) are projections of the same principal circle (i) (and in dotted line) of radius $a$, which can be used to estimate the marker’s observed size and distance from the camera.}
\label{fig:ellipse}
\vspace{-4mm}
\end{figure}

The core of our localisation system involves detecting and tracking a marker comprising LEDs arranged in a circle. The choice of a circle for our proposed marker is inspired by recent works that have explored detection of ellipses for pose estimation tasks \cite{R64,R60}. The perspective projection of a circle onto any 2D plane can be expressed as the equation of an ellipse \cite{R45}. We assume that the observed ellipse is small relative to the full video frame, near the frame's centre, and subject to minimal distortion. This enables us to approximately express an ellipse as the orthogonal projection of its principal circle \cite{R46}, with its semi-major axis length equal to half the diameter of its principal circle. Fig.~\ref{fig:ellipse} illustrates that the solid-lined ellipses in (ii), (iii) and (iv) are projections of the same principal circle (i) from the same position but at different orientations relative to the image plane. We use the observed length of the ellipse’s semi-major axis, $a$, to estimate the LED marker's size in pixels regardless of its orientation.

\begin{figure*}[t]
\centering
\centerline{\includegraphics[width=1.0\textwidth]{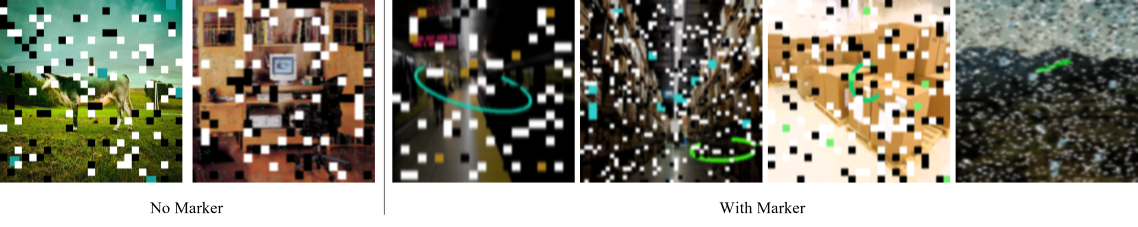}}
\vspace{-1mm}
\caption{Samples of our synthetic data (with and without marker) generated using domain randomisation and other augmentation techniques. The projections of our real circular LED marker are simulated as complete or incomplete ellipses with variations of its parameters.}
\label{fig:dataset}
\vspace{-6mm}
\end{figure*}

\subsection{Generation of Synthetic Dataset for Training CNN}
We propose to use domain randomisation \cite{R41} to simulate our circular LED marker. We simulate green LEDs for training the CNN but a red or blue real LED marker can also be detected by swapping the red or blue channel with the green channel of each input image before each prediction. We generate the outline of an ellipse in each RGB image of 299 x 299 pixels and vary the following, referring to (iv) in Fig.~\ref{fig:ellipse}:
\begin{itemize}
\item A background randomly chosen from Landscape \cite{R47} dataset and MIT Indoor Scenes \cite{R48} dataset.
\item Marker’s shade of green, where \begin{math}R, G, B \in\mathbb{Z}^{+}: G\in[150,255], R\in[0,G-15], B\in[0,G-15]\end{math};
\item Ellipse’s semi-major axis, \begin{math}a\in\mathbb{Z}^{+}: a\in[20,140]\end{math};
\item Ellipse’s semi-minor axis, \begin{math}b\in\mathbb{Z}^{+}:b\in[0,a]\end{math};
\item Position \begin{math}(u, v)\end{math} of the ellipse’s centroid such that the whole ellipse is contained within each image;
\item Ellipse’s rotational angle (in radians), \begin{math}\phi\in[-\pi, +\pi]\end{math};
\item The start angle of the elliptic arc \begin{math}\in[-\pi/2,0]\end{math} and its end angle \begin{math}\in[+\pi,+3\pi/2]\end{math}, to simulate incomplete ellipses when the marker is partially occluded.
\end{itemize}

Each image is then further augmented by varying its brightness, adding different shades of green pixels at random positions, and applying NoisyCutout \cite{R24} to add blur, distractors, and occlusions. We generate a total of 9000 images, with 4500 images containing the ellipse, and the remaining 4500 using the same method but without the ellipse. Samples of our synthetic dataset are shown in Fig.~\ref{fig:dataset}.

\subsection{CNN for Ellipse Detection} \label{cnn}
\textit{1) CNN Architecture:} We propose to leverage the concept of Ellipse R-CNN \cite{R44}, a state-of-the-art CNN-based ellipse detector, to detect a single moving LED marker in frames streamed live from an active camera. While Ellipse R-CNN detects multiple elliptical objects within an image, our approach detects only one projection of our circular LED marker within a small region of interest (ROI) and infer its elliptical parameters. We use MobileNetV2 \cite{R50} (width multiplier of 1.4) for our base CNN architecture as it is computationally efficient and add a classification layer to predict m (the marker’s presence) and a regression layer to predict ellipse parameters $u, v, a, b,$ and $\phi$. We use a sliding window similar to Oh et al. \cite{R24} but with a window size of 448 x 448 pixels as the ROI and an overlap of 50 pixels between windows before resizing this ROI to 299 x 299 x 3 as input into our network.

\textit{2) Loss Function:} Our CNN’s loss function is defined as:

\begin{multline}
\mathcal{L}(I) = \mu_{1}BCE(m,\hat{m}) + \varepsilon[\mu_{2}\frac{\mathcal{L}_c(I)}{d} + \mu_{3}\frac{\mathcal{L}_a(I)}{d^2} \\ + \mu_{4}\frac{\mathcal{L}_b(I)}{d^2} + \mu_{5}\mathcal{L}_\phi(I)]
\label{eq:1}
\end{multline}

where \begin{math}BCE\end{math} is binary cross-entropy for 2-class classification loss, \begin{math}\mathcal{L}_c(I) = \|c-\hat{c}\|_2, \mathcal{L}_a(I) = |a-\hat{a}|, \mathcal{L}_b(I) = |b-\hat{b}|\end{math}, and \begin{math}\mathcal{L}_\phi(I) = |\end{math}atan2\begin{math}[\sin{(\phi-\hat{\phi})},\cos{(\phi-\hat{\phi})}]|\end{math} which rectifies the angular loss (such that for example, the error between $-\pi$ and $+\pi$ is zero and not $2\pi$). $c$ is the predicted pixel coordinates of the ellipse’s centroid $(u,v)$. $\hat{m}, \hat{a}, \hat{b}$ and $\hat{\phi}$ are the predictions of parameters as defined in section \ref{cnn}-1, while $m,c,a,b$ and $\phi$ represent their ground truth. $\mu_i$ are weights to balance the different losses during training and we find these to work well: $\mu_1=0.1, \mu_2=10,\mu_3=10,\mu_4=5$ and $\mu_5=1$. $d$ is the diameter of the detected ellipse's principle circle which is twice of its semi-major axis, $d=2a$, and is used to leverage uncertainty propagation in the losses \cite{R24}. $\varepsilon$ is an object-factor, where $\varepsilon=1$ when the marker is detected and $\varepsilon=0$ otherwise.

\textit{3) Implementation and Training:} Our CNN is implemented in TensorFlow and trained on a Nvidia RTX 2080Ti GPU. During training, pixels of all input images and all ground truths and predictions are normalised to between 0 and 1, except for $\phi$ to between -1 and 1. Our network is optimised using ADAM \cite{R51}, with a batch size of 25, a learn rate of $10^{-4}$ and trained for 200 epochs. We also start training from weights pre-trained on ImageNet \cite{R52} for our MobileNetV2 base architecture. The 9000 generated synthetic images are split into 7000 for training and 2000 for validation.

\subsection{Active PTZ Camera Control for Marker Tracking}
Our method assumes the use of an industrial PTZ camera for its accessible API, precise pan-tilt-zoom control, as well as its minimal distortion in video frames and we use a Panasonic AW-HE40H PTZ camera to demonstrate our method. This allows us to only require calibrating its FOV at different levels of zoom. However, the control of such cameras is often limited by their control latency, which is about 130ms for our camera. To reduce processing time required for sending commands to the PTZ camera to query its zoom, and the tedious or impractical process of calibrating for the entire zoom range (for example, our camera has up to 2730 zoom steps), we propose to use pre-defined zoom levels. Auto-focus is disabled and set to its furthest setting to maximise FOV and prevent uncontrolled changes during operation. The camera’s brightness is also set to its lowest level to reduce exposure to other light sources and help the LED marker appear clearer. In our work, we choose to split our camera’s zoom into eight discrete levels and perform camera calibration to obtain their HFOV: 54{\degree}, 45.01{\degree}, 37.87{\degree}, 30.67{\degree}, 24.21{\degree}, 18.11{\degree}, 12.99{\degree} and 8.59{\degree}, assigned to zoom states 0 to 7 respectively.

\textit{1) Zoom Control:} We control the camera’s zoom with the objective of maintaining the size of the observed marker's diameter within the ROI. We use a simple algorithm for this control, where we start from zoom state 0 (fully zoomed out) and increase the zoom state by 1 (upper limit of state 7) when the marker is observed to be smaller than $\frac{1}{4}$ of the ROI’s width and decrease it by 1 (lower limit of state 0) when larger than $\frac{7}{12}$ of the ROI’s width. An interval of one second between marker size checks is set to prevent erratic zoom behaviour.

\textit{2) Pan-tilt Control:} To track the detected marker, we compute and send pan and tilt commands to the PTZ camera proportional to the marker’s angle from the image centre. The following equations are used to relate the pixel difference from the image’s centre $(\Delta{u}, \Delta{v})$ to pan and tilt angles $(\theta_p, \theta_t)$ as a function of the camera’s horizontal (H) and vertical (V) FOV:

\begin{equation}
\begin{split}
\theta_p &= \arctan{[\Delta{v}*2\tan{(0.5*HFOV)}]} \\
\theta_t &= \arctan{[\Delta{u}*2\tan{(0.5*VFOV)}].}
\label{eq:2}
\end{split}
\end{equation}

For every processed video frame, pan and tilt commands $(\mu_{pan}, \mu_{tilt})$ are only sent to the PTZ camera if pan or tilt angle differences exceeds 1, as described by the following:

\begin{equation}
\begin{split}
\mu_{pan} = \frac{\theta_p}{s}, & \quad\text{if }\theta_p>1\\
\mu_{tilt} = \frac{\theta_t}{s}, & \quad\text{if }\theta_t>1.
\label{eq:3}
\end{split}
\end{equation}

We use a scaling factor, $s = 4$, to reduce the chance of over-correction and find this to work in all our experiments.

\subsection{Leveraging Elliptical Angle to Improve Range Estimation}
\textit{1) Angle of Ellipse as a Measure of Uncertainty:} Our approach assumes that common robotic platforms including wheeled robots or slow-moving UAVs used in applications such as building inspection strives to maintain its base parallel to the ground plane. This implies that our proposed circular LED marker, designed to be attached such that its plane is parallel to the robot's base, should be mostly observed as an ellipse with a relatively horizontal major axis (having a small angle, $|\phi|$) within the camera frame. Conversely, a large $|\phi|$ may suggest a more dynamic state (such as a fast moving UAV with a large roll) or an inaccurate visual prediction. Our experimental data in Table~\ref{tab_corr} also supports this, showing a positive correlation between $|\phi|$ and the error in estimated marker distance from the camera. Hence, we propose to use $|\phi|$ as a method to measure the uncertainty of predictions. 

\textit{2) Leveraging $|\phi|$ in a Filter:} Due to its adaptability and ease of implementation, we choose to use a SIR particle filter \cite{R57} for state estimation and to demonstrate one method of leveraging $|\phi|$ to improve localisation performance. We only use the filter to update the estimated distance of the marker from the camera, $\rho$, as we find it does not improve the remaining two angular estimates in the Spherical Coordinate System. We obtain $\rho$ using this equation:

\begin{equation}
\rho = \frac{Dh}{2d_o}\cot{(0.5*HFOV)}.
\label{eq:5}
\end{equation}

Where $D$ is the actual metric diameter of the circular LED marker, $h$ is the horizontal pixel resolution of the camera, and $d_o$ is the observed diameter of the marker in pixels within the ROI image, which is obtained using $d_o = 2a$.

\setlength{\textfloatsep}{2pt}
\begin{algorithm}[t]
\caption{Summary of Proposed Adaptive Particle Filter}
\begin{algorithmic}[1]

\STATE\textbf{Initialisation: } Generate initial sample set using Normal distribution 
\STATE\textbf{Prediction: } Draw predicted samples; N = 2000 particles, using linear dynamics function: $\rho_k$ = $\dot{\rho}_{k-1}dt$ and Gaussian noise with $\sigma_\rho$ = 0.3 and $\sigma_{\dot{\rho}}$ = 0.1
\STATE\textbf{Update: } Use observation data $\rho_k$ and predicted samples;

Compute particles weights based on Equations~\ref{eq:rbf} and \ref{eq:sigmarbf} (proposed use of $|\phi|$ for adaptive $\sigma_{rbf,k}$)
\STATE Normalise particle weights and use sum of all weighted particles as current state estimate
\STATE\textbf{Resampling: } Systematic resampling \cite{R58} of particles
\STATE Go back to step 2
\end{algorithmic}
\label{alg:apf}
\end{algorithm}

We summarise our adaptive particle filter in Algorithm~\ref{alg:apf}. In particular, we propose to use the magnitude of the predicted angle $|\phi|$ to adapt the standard deviation term, $\sigma_{rbf}$ used in the radial basis function (RBF) for computing particle weights during the update step, in the following equation:

\begin{equation}
RBF(\rho_k,\mathrm{x}_k^i) = \exp(-\frac{\|\rho_k-\mathrm{x}_k^i\|^2}{2\sigma_{rbf,k}^2}).
\label{eq:rbf}
\end{equation}

Where $\mathrm{x}_k^i$ represents each particle at time $k$ and:

\begin{equation}
\sigma_{rbf,k} = \max(\lambda|\phi_k|,\sigma_{rbf,min}).
\label{eq:sigmarbf}
\end{equation}

Where $\lambda$ is a fixed scalar experimentally determined to be 15 and $\sigma_{rbf,min}$ is set at 0.5 to prevent assigning too high weights to particles very near to the observation when $\phi$ is near zero. Equations~\ref{eq:rbf} and \ref{eq:sigmarbf} function such that as $|\phi|$ (which we propose to use as a measure of prediction uncertainty) increases, $\sigma_{rbf,k}$ also increases to reduce the weights of particles that are nearer to the observation so that it has lesser influence on the estimate. 

We obtain the 3D Cartesian coordinates of a detected marker by converting from spherical coordinates using the following equations (accounting for PTZ camera’s current pan and tilt queried at every frame as well as $\theta_p$ and $\theta_t$):

\begin{equation}
\begin{split}
x &= \rho\sin{(tilt+\theta_t)}\\
y &= \rho\cos{(tilt+\theta_t)}\sin{(pan+\theta_p)}\\
z &= \rho\cos{(tilt+\theta_t)}\cos{(pan+\theta_p)}
\label{eq:4}
\end{split}
\end{equation}

We find that a 3rd order Butterworth filter (with 0.6 Hz critical frequency) is effective in smoothing HFOV values when transitioning between zoom states before computing $\rho$, and use a 1st order Butterworth filter (2 Hz critical frequency) to smoothen all $(tilt+\theta_t)$ and $(pan+\theta_p)$ outputs.

\begin{figure}[t]
\centerline{\includegraphics[width=1.0\columnwidth]{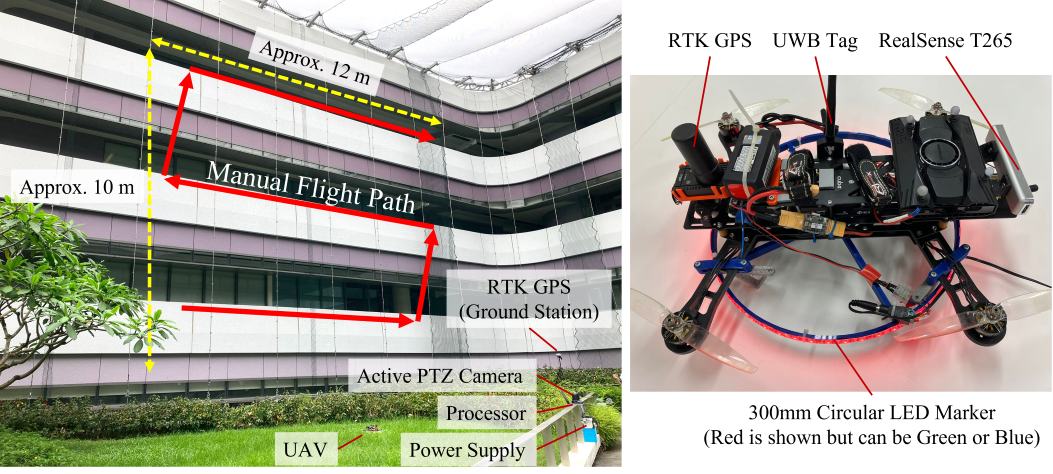}}
\vspace{-1mm}
\caption{\textit{Left}: Experimental setup simulating a possible flight profile for building inspection as a potential application. \textit{Right}: The UAV used in our experiments attached with our proposed circular LED marker to its underside. RTK GPS is used for ground truth, with a RealSense T265 and UWB for comparison.}
\label{fig:arena}
\vspace{-3mm}
\end{figure}

\section{Experiment and Results}
We build a circular LED marker of diameter $D$ = 0.30 m using an LED strip curved around a 3D printed circular fixture. We conduct real-world experiments using a UAV, with the LED marker mounted to its underside, as the target robotic platform to evaluate our proposed localisation system and demonstrate its capability. The Panasonic AW-HE40H PTZ camera used in our experiments has a frame rate of 12.5 fps at 1920X1080 resolution, pan-tilt steps of approximately 0.02{\degree}, and a control latency of about 130 ms. While the average CNN inference time is about 23ms with a Nvidia RTX 1650Ti GPU, the control latency limits our system to about 8 Hz but this may be improved with better hardware or implementation.

\begin{figure}[t]
\centerline{\includegraphics[width=1.0\columnwidth]{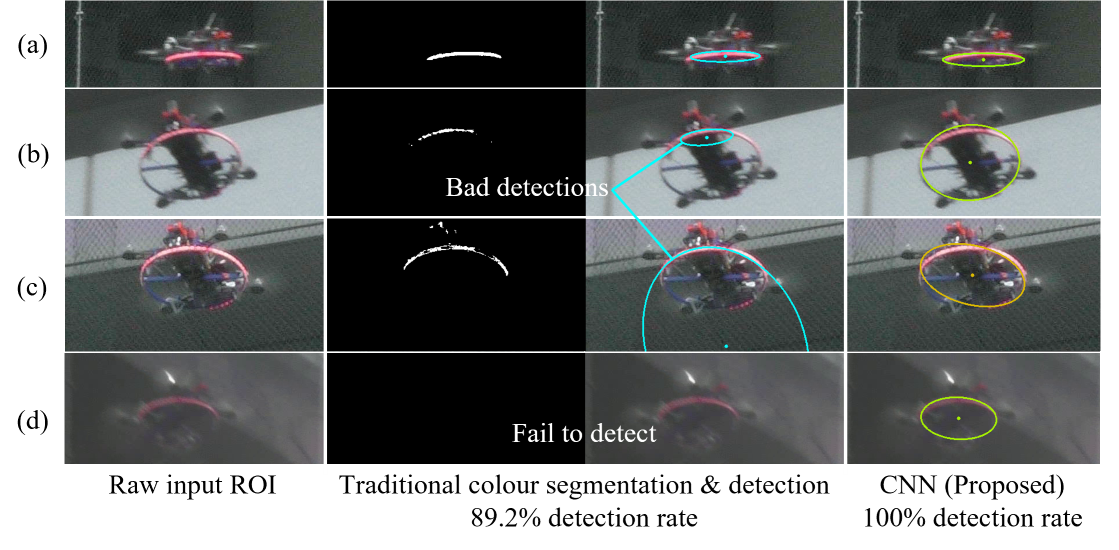}}
\vspace{-3mm}
\caption{Detections of our circular LED marker in an unstructured environment using traditional ellipse detection \cite{R63} and our proposed CNN-based approach. Our approach is more robust in detecting and estimating the size of the LED marker without requiring tuning of parameters for feature extraction.}
\label{fig:trad_vs_ours}
\vspace{-1mm}
\end{figure}

\subsection{Comparing Robustness With Traditional Ellipse Detection} \label{robustness}
We conduct a flight experiment using our UAV within an outdoor netted arena designed for testing drones. The UAV is manually piloted (at approximately 0.5 to 1.5 ms\textsuperscript{-1}) along a vertical 'S' path while near and facing one side of the building to simulate a possible flight profile for building inspection as a potential application, while the PTZ camera actively tracks the LED marker. The setup for this experiment is shown in Fig.~\ref{fig:arena} with the UAV used for all our experiments.

Fig.~\ref{fig:trad_vs_ours} shows that our proposed CNN-based approach is more robust in detecting the LED marker as compared to using traditional ellipse detection (TED) \cite{R63}. TED only has a 89.2\% detection rate (fail to detect 59 out of 548 frames) while our CNN-based approach constantly detects the marker throughout the experiment. Table~\ref{tab:apf} and Table~\ref{tab:3dloc} also show that using TED results in substantially larger errors due to frequent inaccurate marker detection. Furthermore, our approach does not require parameter tuning for feature extraction.

\begin{figure}[t]
\centerline{\includegraphics[width=1.0\columnwidth]{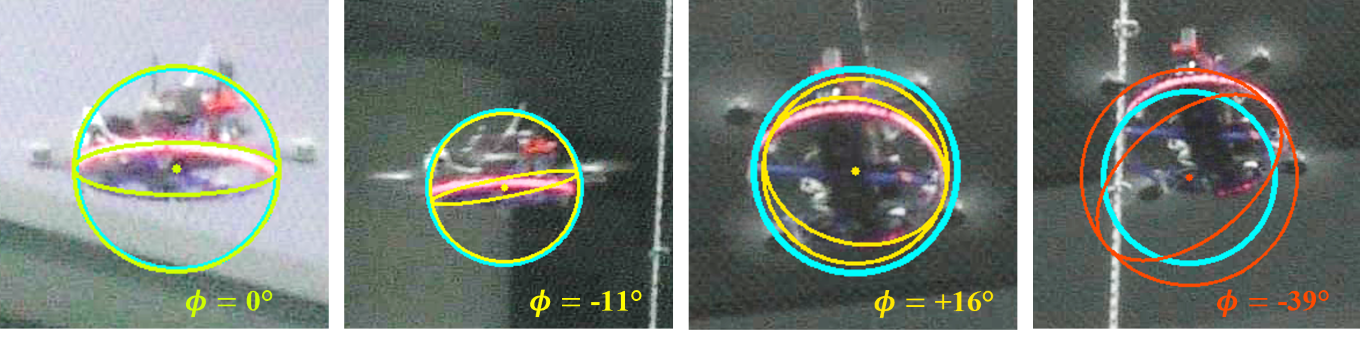}}
\vspace{-1mm}
\caption{\textit{Left to right}: Examples with increasing predicted ellipse angle, $|\phi|$ and increasing discrepancy in predicted marker size (from green to red, with increasing redness representing a larger $|\phi|$ value). This shows the viability of using $|\phi|$ as a measure of prediction uncertainty. Blue circles represent size estimates derived from improved range estimates using our proposed APF.}
\label{fig:angle_error_eg}
\vspace{-2mm}
\end{figure}


\captionsetup[table]{position=bottom}

\newcolumntype{C}{>{\centering\arraybackslash}X}
\newcolumntype{R}{>{\raggedright\arraybackslash}X}

\begin{table}[t]
\vspace{+0mm}
\begin{center}
\begin{tabularx}{1.0\columnwidth}{C C}
\hline
Pearson Correlation Coefficient & +0.804 \\
\hline
Spearman Correlation Coefficient & +0.352 \\
\hline

\end{tabularx}
\end{center}
\vspace{-3mm}
\caption{Correlation between the magnitude of ellipse angle, $|\phi| > 0.175$, and the error in predicted marker distance from the camera before filtering.}
\vspace{-0mm}
\label{tab_corr}
\end{table}

\subsection{Leveraging Elliptical Angle to Improve Range Estimation} \label{apf}
Using the same flight experiment described in section \ref{robustness}, we evaluate our proposed use of the predicted elliptical angle $\phi$ as a measure of prediction uncertainty and leveraging $|\phi|$ in an adaptive particle filter to improve range estimation. A total of 137 data points from the experiment are compared with their ground truth obtained using RTK GPS at about 2 Hz.

\begin{figure}[h!]
\centerline{\includegraphics[width=1.0\columnwidth]{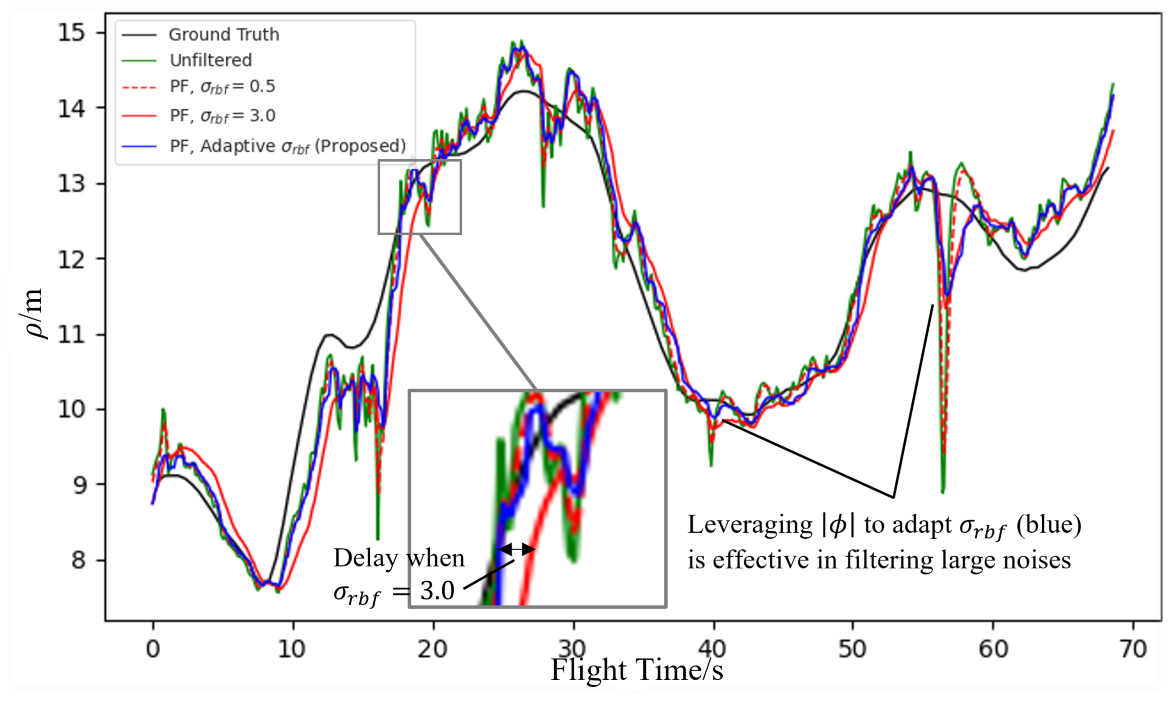}}
\vspace{-3mm}
\caption{Our proposed use of the predicted elliptical angle $|\phi|$ to adapt $\sigma_{rbf}$ in a particle filter is effective in filtering large noises in range ($\rho$) predictions. In comparison, using a fixed value of $\sigma_{rbf}$=0.5 only achieves minor improvements while using $\sigma_{rbf}$=3.0 introduces a substantial delay.}
\label{fig:rhoplot}
\vspace{-1mm}
\end{figure}

\textit{1) Angle of Ellipse as a Measure of Uncertainty:} Fig~\ref{fig:angle_error_eg} shows examples from the experiment where a larger magnitude of predicted angle $|\phi|$ generally corresponds to a larger discrepancy between the predicted and actual size of the marker. We also observe a positive correlation (see Table~\ref{tab_corr}) between data points with $|\phi| > 0.175$ (corresponding to $10\degree$) and the error in range estimates. While the value of $|\phi|$ is not always proportional to prediction error, our results show the viability of using $|\phi|$ as a measure of uncertainty in size predictions. 

\textit{2) Improvement in Range Estimates:} Fig.~\ref{fig:rhoplot} shows $\rho$ estimates using different $\sigma_{rbf}$. We observe that our proposed use of $|\phi|$ to adapt $\sigma_{rbf}$ is effective in filtering large noises in $\rho$ predictions. In comparison, using a fixed small $\sigma_{rbf}$ of 0.5 only achieves minor smoothing while using a fixed larger value of 3.0 introduces a substantial delay. Quantitative results in Table~\ref{tab:apf} show that our method substantially reduces RMSE of $\rho$ estimates without considerably increasing median error. This is as opposed to using a small $\sigma_{rbf}$ (0.5) that only slightly improves RMSE or a larger $\sigma_{rbf}$ (3.0) that results in a higher RMSE due to the increase in delay of state estimates. Blue circles in Fig.~\ref{fig:angle_error_eg} show how improved range estimation using our proposed method also improves size estimation in instances where $|\phi|$ is large without affecting good predictions where $|\phi|$ is small. Hence, we conclude that leveraging $|\phi|$ as a measure of uncertainty and using it in an adaptive particle filter is effective in improving range estimates of our marker.


\captionsetup[table]{position=bottom}

\newcolumntype{C}{>{\centering\arraybackslash}X}
\newcolumntype{R}{>{\raggedright\arraybackslash}X}

\begin{table}[t]
\vspace{+0mm}
\begin{center}
\begin{tabularx}{1.0\columnwidth}{m{4.3cm} m{1.6cm} m{1.8cm}}
\hline
\centering Filter (for $\rho$) & \centering Median & \centering\arraybackslash RMSE \\
\hline
Trad. Ellipse Det. \cite{R63}, BW
 & \centering {0.895 m} & \centering\arraybackslash 6.884 m \\
\hline
CNN, No Filter & \centering\boldblue{0.238 m} & \centering\arraybackslash 0.528 m \\
\hline
CNN, BW, $f_{crit}$=1Hz & \centering 0.244 m & \centering\arraybackslash 0.524 m\\
\hline
CNN, PF, $\sigma_{rbf}$=0.5 & \centering 0.244 m & \centering\arraybackslash 0.517 m \\
\hline
CNN, PF, $\sigma_{rbf}$=3.0 & \centering 0.341 m & \centering\arraybackslash 0.577 m\\
\hline
CNN, APF using $|\phi|$ (Proposed) & \centering 0.239 m & \centering\arraybackslash\boldblue{0.458 m} \\
\hline

\end{tabularx}
\end{center}
\vspace{-3mm}
\caption{Errors in estimation of $\rho$, comparing our proposed use of CNN-based detection while using $|\phi|$ in an adaptive particle filter (APF) against using traditional ellipse detection \cite{R63} and other non-adaptive filters. BW: $1^{st}$ Order Butterworth filter with $f_{crit}$=1Hz.}
\vspace{-1mm}
\label{tab:apf}
\end{table}

\subsection{3D Localisation Accuracy}
We compare 3D localisation accuracy of our proposed method against common alternatives for localisation in GPS-denied environments: a) VIO using an onboard RealSense T265 (20Hz) and b) an onboard Nooploop LinkTrack UWB tag (50Hz) with four UWB on-ground anchors. We collect data from these sensors in the same experiment described in section \ref{robustness} for comparison. Results in Table~\ref{tab:3dloc} show that our proposed method leveraging $|\phi|$ in an adaptive particle filter achieves the lowest error. Additionally, the errors in Table~\ref{tab:3dloc} are very similar to their respective errors in Table~\ref{tab:apf}, showing that $\rho$ estimates contribute largest to 3D localisation error.


\captionsetup[table]{position=bottom}

\newcolumntype{C}{>{\centering\arraybackslash}X}
\newcolumntype{R}{>{\raggedright\arraybackslash}X}

\begin{table}[t]
\vspace{+0mm}
\begin{center}
\begin{tabularx}{1.0\columnwidth}{m{4.8cm} m{1.6cm} m{1cm}}
\hline
\centering 3D Localisation Method & \centering Median & \centering\arraybackslash RMSE \\
\hline
UWB [4 Anchors/1 Tag]+BW for XYZ & \centering 1.179 m & \centering\arraybackslash 1.446 m\\
\hline
Intel RealSense T265+no added filter & \centering 0.547 m & \centering\arraybackslash 0.610 m \\
\hline
Marker+Trad. Ellipse Det. \cite{R63}+BW for $\rho$ & \centering 0.905 m & \centering\arraybackslash 6.886 m \\
\hline
Marker+CNN+No added filter (Proposed) & \centering 0.283 m & \centering\arraybackslash 0.544 m \\
\hline
Marker+CNN+BW for $\rho$ (Proposed) & \centering 0.287 m & \centering\arraybackslash 0.540 m\\
\hline
Marker+CNN+APF using $|\phi|$ (Proposed) & \centering\boldblue{0.277 m} & \centering\arraybackslash\boldblue{0.476 m} \\
\hline

\end{tabularx}
\end{center}
\vspace{-3mm}
\caption{3D localisation median error and root-mean-square-error (RMSE)}
\vspace{-1mm}
\label{tab:3dloc}
\end{table}

\begin{figure}[t]
\centerline{\includegraphics[width=1.0\columnwidth]{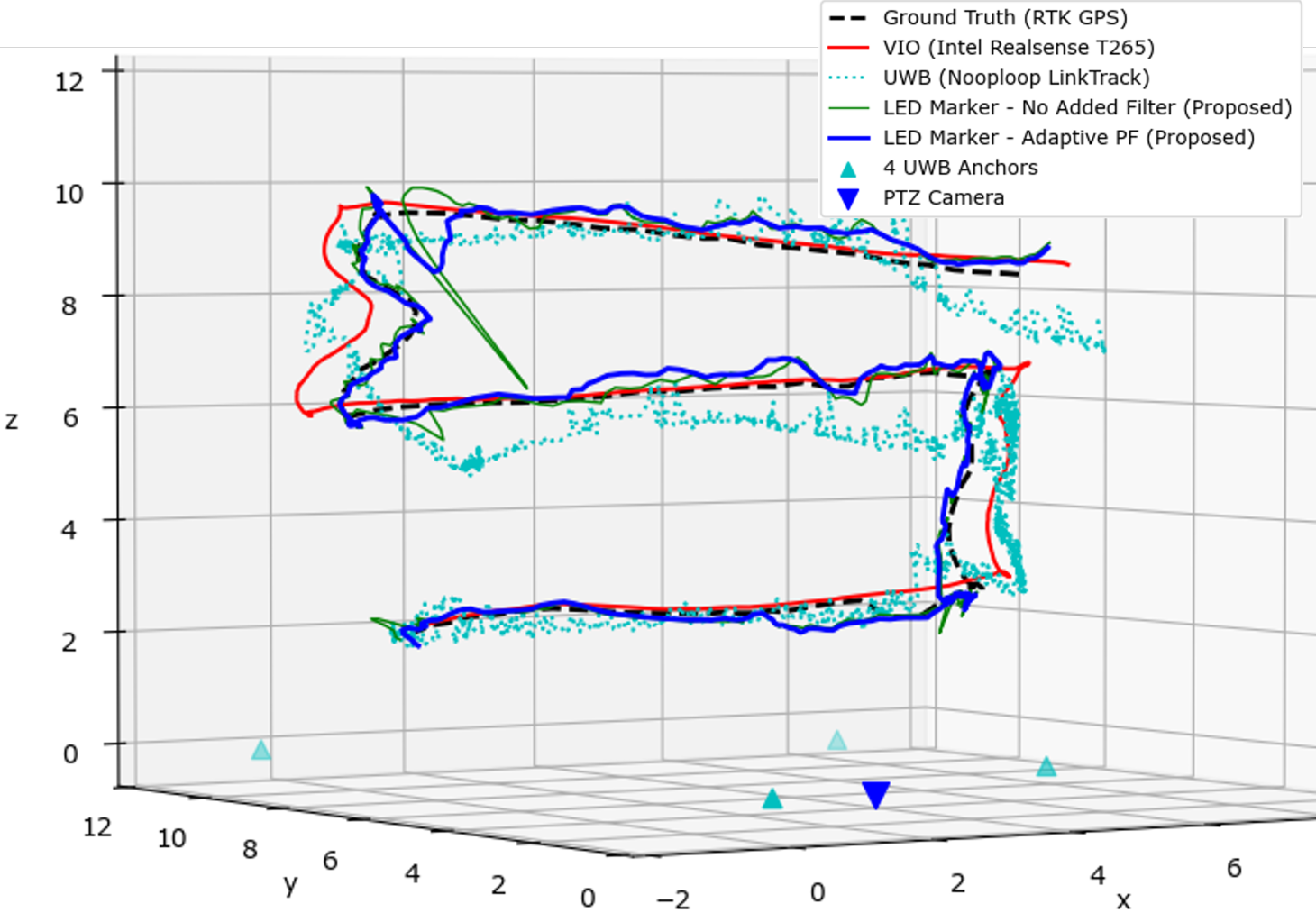}}
\vspace{-1mm}
\caption{3D localisation plot. Our proposed method leveraging $|\phi|$ in an adaptive particle filter successfully tracks the LED marker along the entire flight path and achieves the best localisation accuracy compared to the other approaches. Additionally, our method does not suffer from drift unlike using VIO, and requires no on-site calibration and fewer infrastructure as compared to UWB.}
\label{fig:3dloc}
\vspace{-3mm}
\end{figure}

\begin{figure}[h!]
\centerline{\includegraphics[width=1.0\columnwidth]{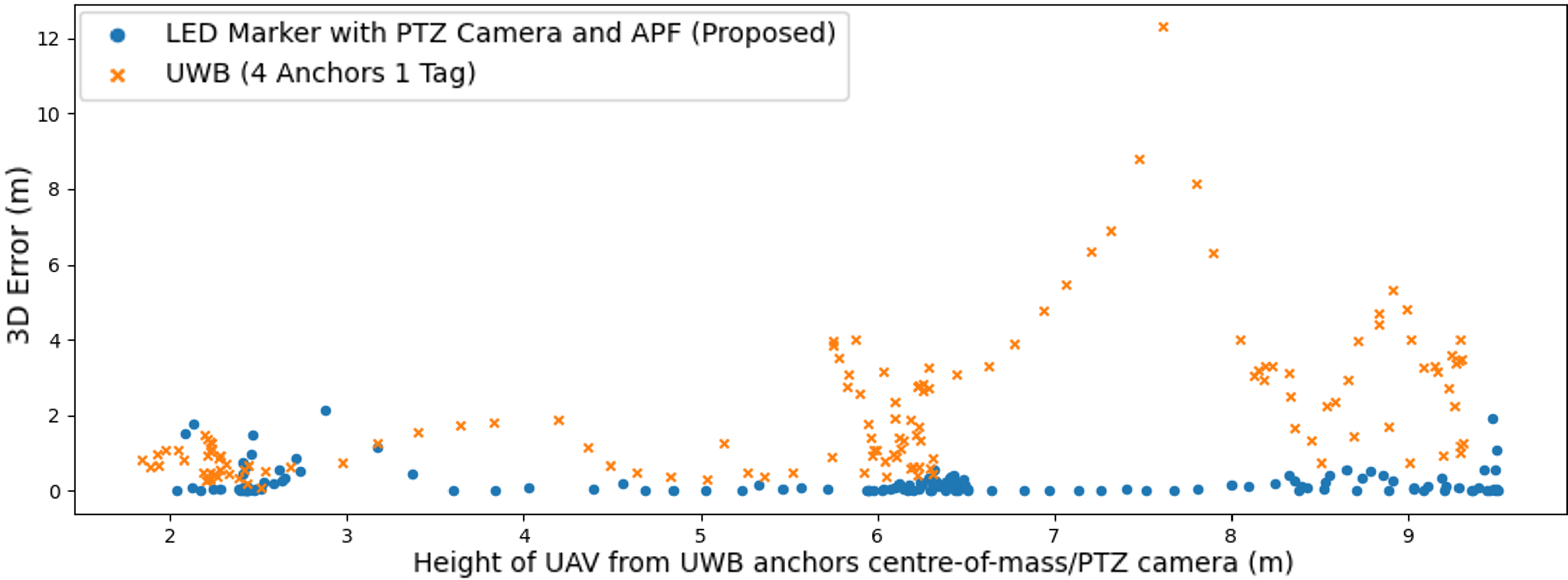}}
\vspace{-1mm}
\caption{3D localisation error of the UWB system (using 4 anchors in a rectangular configuration along the same plane) increases more with height from the centre of mass of its anchors as compared to the error of our proposed LED marker-based system with increasing height from the PTZ camera.}
\label{fig:proposedvsuwb}
\vspace{-1mm}
\end{figure}

Qualitatively, we can observe in Fig.~\ref{fig:3dloc} that our proposed approach of tracking and localising a circular LED marker using an active PTZ camera outperforms the other methods in 3D localisation accuracy. Our approach also does not suffer from drift, does not require on-site calibration and has low infrastructural requirements. 

\begin{figure}[t]
\centerline{\includegraphics[width=0.8\columnwidth]{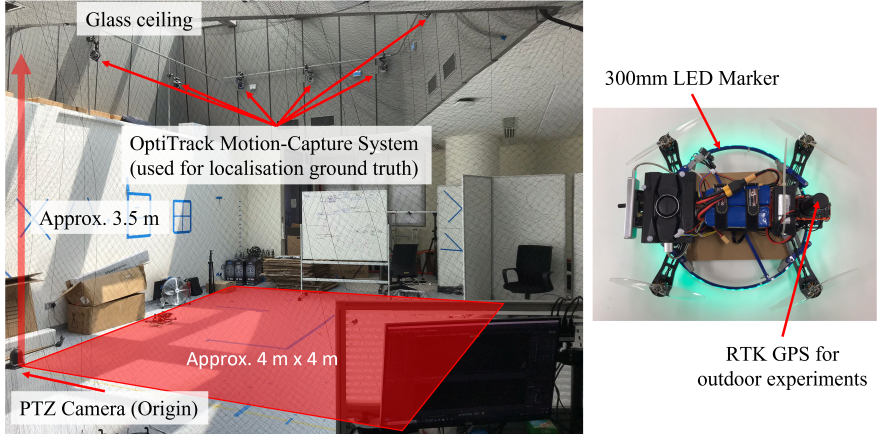}}
\vspace{-1mm}
\caption{\textit{Left}: Indoor experimental setup with a glass ceiling that allows sunlight to enter. A PTZ camera is at a fixed position on the ground (left corner) for tracking and localisation of our LED marker and OptiTrack is used for indoor ground truth. \textit{Right}: The UAV used in our flight experiments and RTK GPS for outdoor ground truth.}
\label{fig:indoor}
\vspace{-3mm}
\end{figure}

\begin{figure}[h!]
\centerline{\includegraphics[width=0.8\columnwidth]{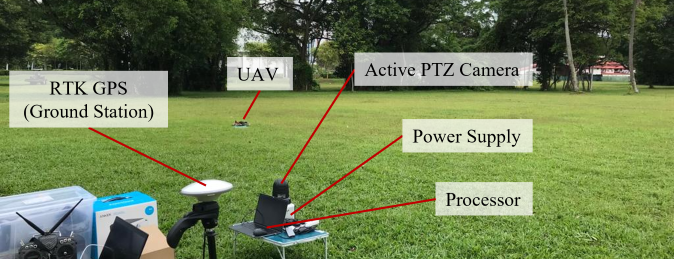}}
\vspace{-1mm}
\caption{Outdoor experimental setup, using the same UAV shown in Fig.~\ref{fig:indoor}. RTK GPS is used for waypoint navigation and ground truth.}
\label{fig:outdoor}
\vspace{-1mm}
\end{figure}

Fig.~\ref{fig:proposedvsuwb} also shows that the error of our proposed localisation system does not substantially increase with increasing marker height from the PTZ camera. On the other hand, the localisation error of the UWB system used in our experiment increases substantially with increasing distance of the tag from the centre of mass of the UWB anchors. This is likely due to the benefit of the camera's ability to zoom and maintain the resolution of the observed marker and demonstrates another potential advantage of our proposed system.

\subsection{Performance in Different Environments}
\begin{figure}[t]
\centerline{\includegraphics[width=1.0\columnwidth]{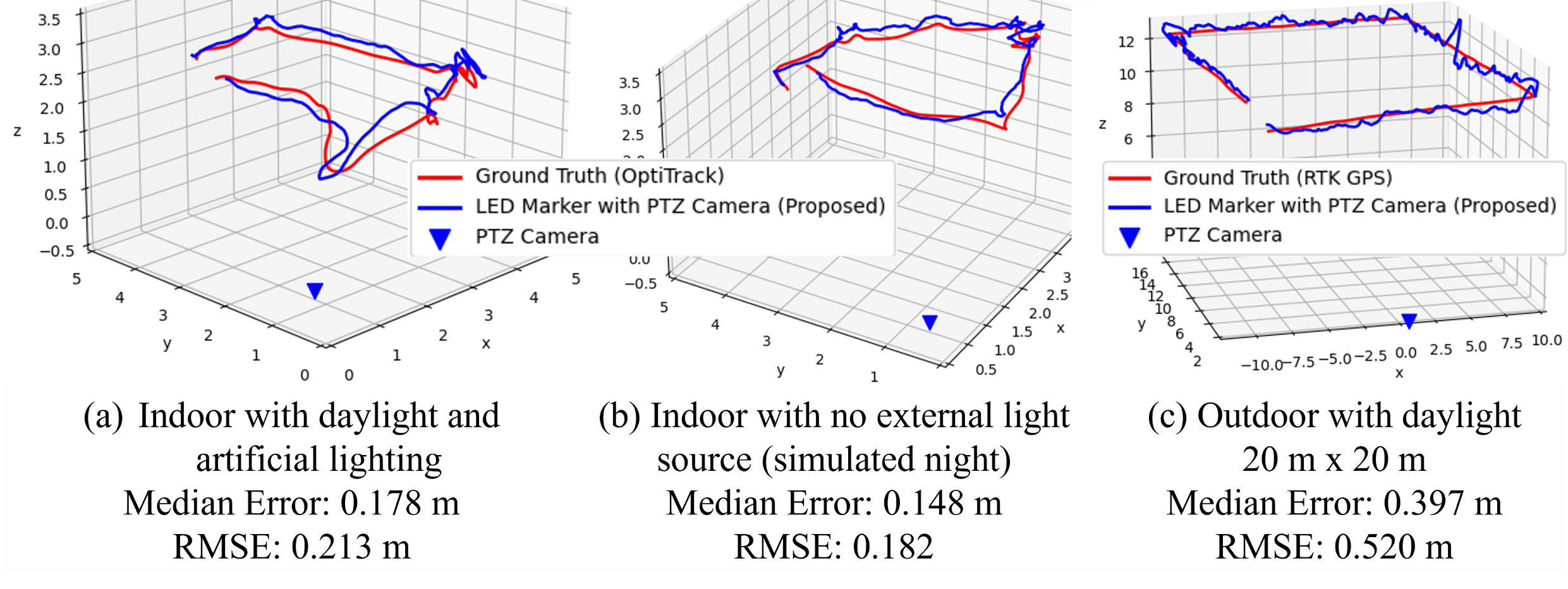}}
\vspace{-1mm}
\caption{Localisation results from three experiments with an approximately square flight profile, including their respective 3D error. These results demonstrate successful tracking and localisation of the UAV in different environments using our proposed localisation system.}
\label{fig:squareplots}
\vspace{-3mm}
\end{figure}

\begin{figure}[t]
\centerline{\includegraphics[width=1.0\columnwidth]{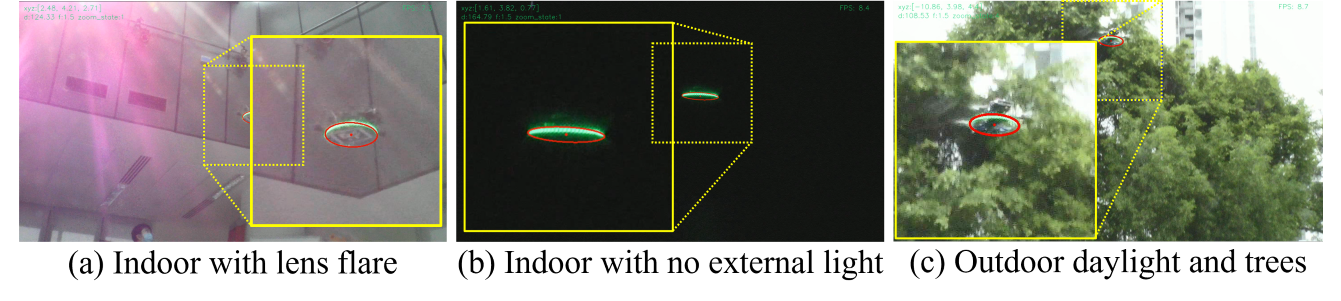}}
\vspace{-1mm}
\caption{Examples of camera frames from different experiments with their ROI (dotted yellow square) enlarged, demonstrating that our proposed LED marker can be robustly detected (outlined in red ellipses) by our CNN.}
\label{fig:robust}
\vspace{-1mm}
\end{figure}

We conduct three additional experiments to demonstrate the performance of our proposed localisation system in indoor and outdoor environments. In each experiment, the UAV travels an approximately square flight profile with the following conditions: (a) Indoor manual flight within a 4 x 4 x 3.5 m$^3$ space with a mix of natural and artificial lighting; (b) Indoor manual flight in the same space without any external light source (also simulating night); (c) Outdoor flight covering 20 m x 20 m (one large square flight profile at about 10 m height) via waypoint navigation using RTK GPS. We use a green LED and swap the green and red channels of each input image into our CNN to demonstrate that our method can work with other colours of LED. The experimental setups are shown in Fig.~\ref{fig:indoor} and Fig.~\ref{fig:outdoor}. We present 3D localisation plots in Fig.~\ref{fig:squareplots}, showing that our method is able to track the green LED marker throughout the UAV's flight path in all three experiments while Fig.~\ref{fig:robust} contains examples of detection of the green LED marker in different scenes. Our results demonstrate robust detection of our proposed marker and successful localisation of the UAV in different environments using our proposed localisation system. 

\section{Conclusion}
This work proposes an easily deployable 3D localisation system that can be used in GPS-denied environments, with minimal infrastructure requirements and not requiring on-site calibration. This is achieved using a single PTZ camera to detect, track, and localise a circular LED marker that can be attached to any target platform. We show that our CNN more robustly detects and predicts the marker's elliptical parameters compared to traditional ellipse detection and does not require parameter tuning for feature extraction. We show that the predicted elliptical angle can be used as a measure of prediction uncertainty and can be leveraged to filter large noises in range estimates and improve 3D localisation accuracy. Results from an experiment that mimics a potential application show that our method achieves better localisation accuracy as compared to alternative solutions for GPS-denied environments. Lastly, we demonstrate the performance of our system in both outdoor and indoor environments with different lighting conditions. Future work can explore modelling the system to improve PTZ camera control or fusion with other sensor data to improve tracking and localisation performance.

\bibliographystyle{./bibliography/IEEEtran}
\bibliography{./bibliography/main}

\begin{IEEEbiography}[{\includegraphics[width=1in,height=1.25in,clip,keepaspectratio]{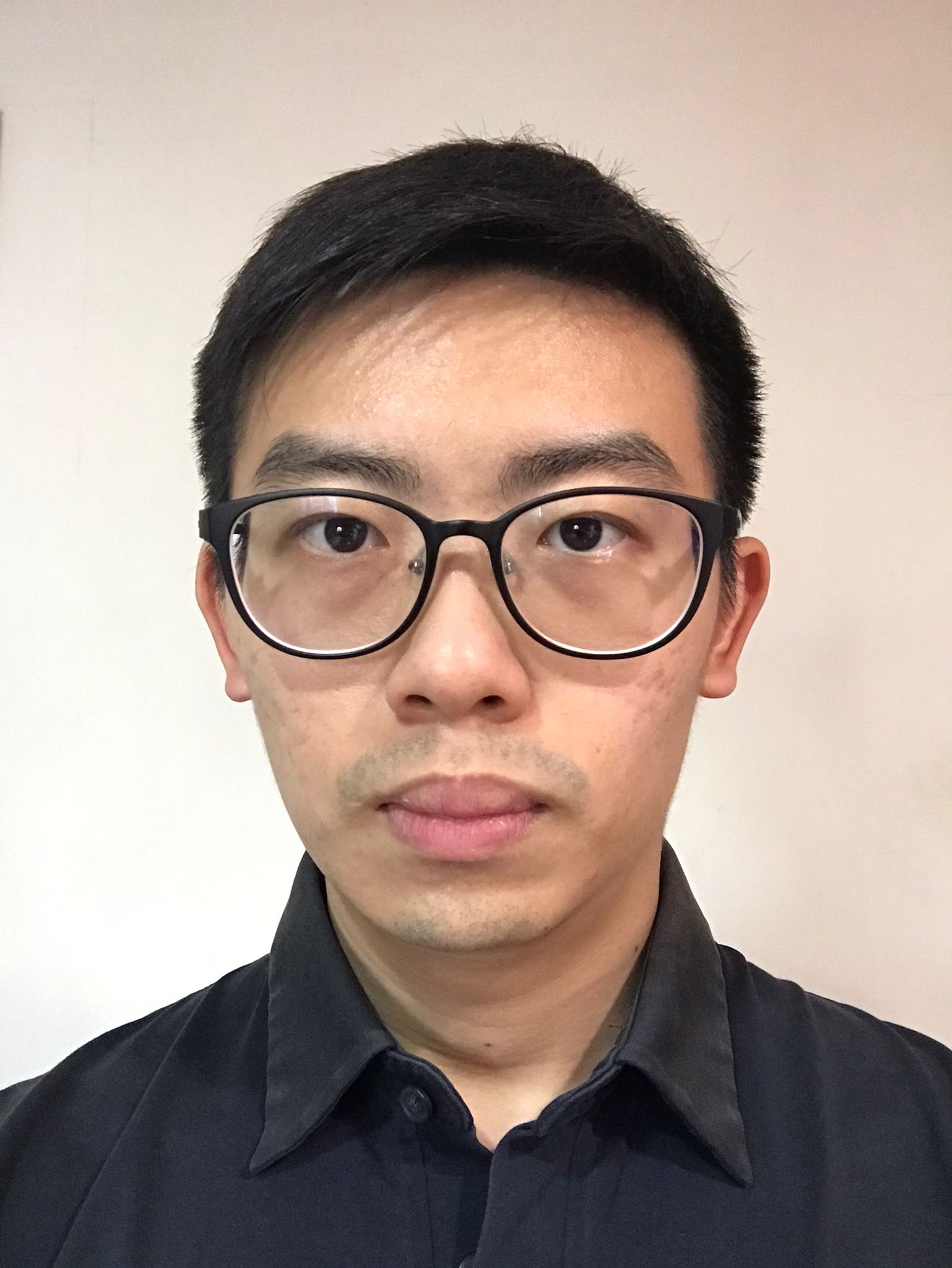}}]{Xueyan Oh}
received the B.Eng. (Hons.) degree in engineering from the Singapore University of Technology and Design (SUTD), Singapore, in 2016. He is currently working toward the Ph.D. degree in Engineering Product Development (EPD) with the SUTD Engineering Product Development (EPD) Pillar, Singapore. His research interests include deep learning and vision-based localisation.
\end{IEEEbiography}

\begin{IEEEbiography}[{\includegraphics[width=1in,height=1.25in,clip,keepaspectratio]{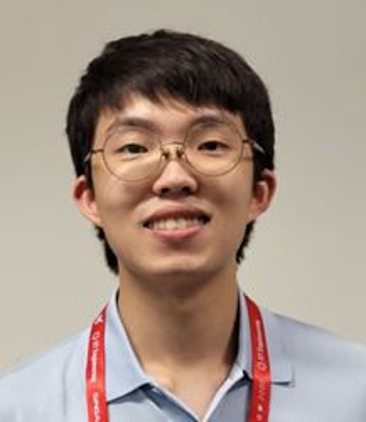}}]{Ryan Lim}
is currently working toward the Ph.D. degree in Engineering Product Development (EPD) at the Singapore University of Technology and Design (SUTD) under the Engineering Product Development pillar. He is with the SUTD Aerial Innovation Robotics Lab (AIRLAB) and his research focuses on the development and integration of novel aerial manipulation mechanisms for unmanned systems, geared towards payload deployment.
\end{IEEEbiography}

\begin{IEEEbiography}[{\includegraphics[width=1in,height=1.25in,clip,keepaspectratio]{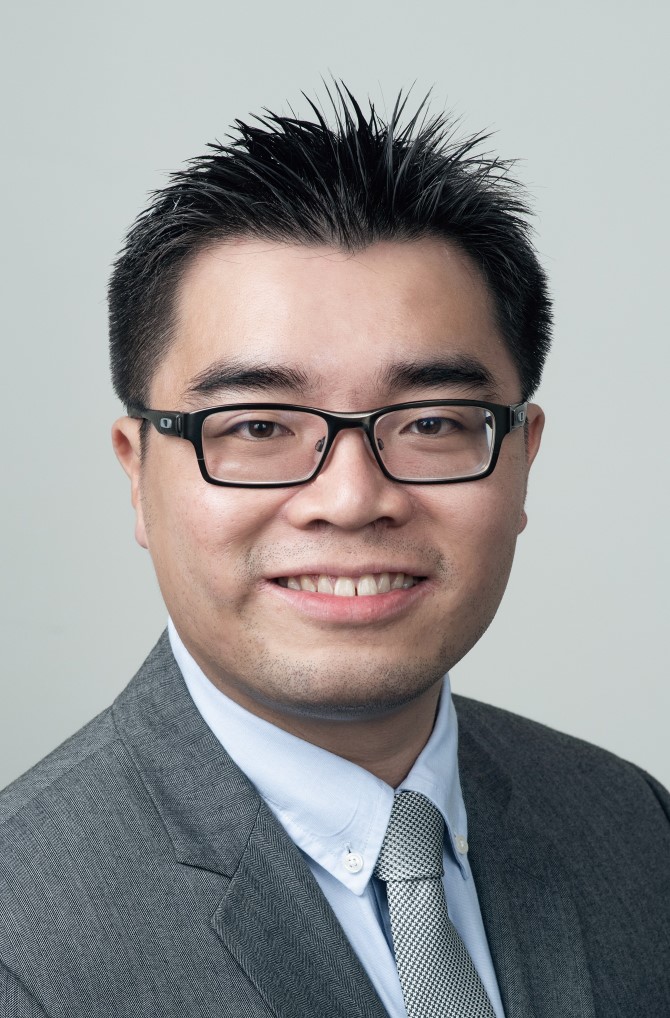}}]{Shaohui Foong}
is an Associate Professor and also the Associate Head of Pillar in the Engineering Product Development (EPD) Pillar at the Singapore University of Technology and Design (SUTD) and Senior Visiting Academician at the Changi General Hospital, Singapore. He received his B.S., M.S. and Ph.D. degrees in Mechanical Engineering from the George W. Woodruff School of Mechanical Engineering, Georgia Institute of Technology, Atlanta, USA, in 2005, 2008 and 2010 respectively. In 2011, he was a Visiting Assistant Professor at the Massachusetts Institute of Technology, Cambridge, USA. His research interests include system dynamics \& control, nature-inspired robotics, magnetic localization, medical devices and design education \& pedagogy.
\end{IEEEbiography}

\begin{IEEEbiography}[{\includegraphics[width=1in,height=1.25in,clip,keepaspectratio]{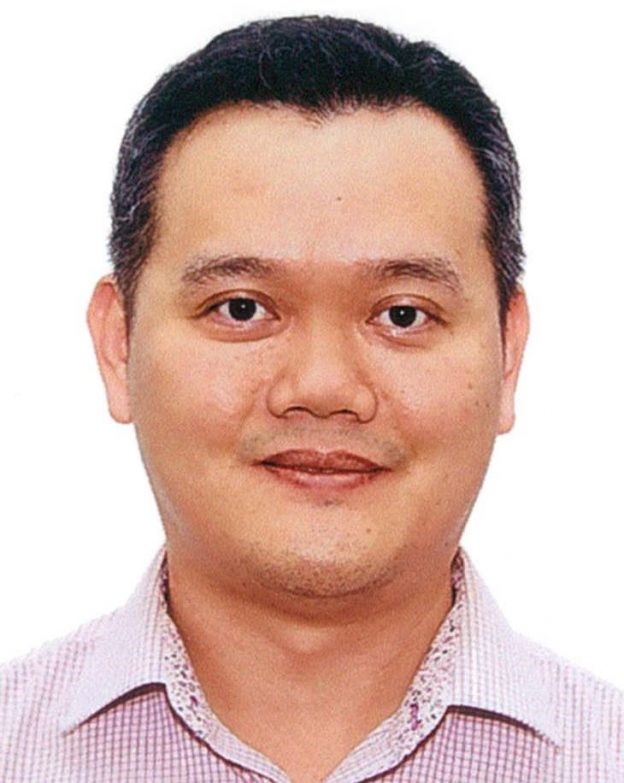}}]{U-Xuan Tan}
(Member, IEEE) received the B.Eng. and Ph.D. degrees from Nanyang Technological University, Singapore, in 2005 and 2010, respectively. From 2009 to 2011, he was a Postdoctoral Fellow with the University of Maryland, College Park, MD, USA. From 2012 to 2014, he was a Lecturer with the Singapore University of Technology and Design (SUTD), Singapore, where he took up a research intensive role in 2014 and has been promoted to Associate Professor since 2021. He is the current Chair of SUTD Institutional Review Board and is also holding a Senior Visiting Academician position at Changi General Hospital. His research interests include mechatronics, on-site robotics algorithm, sensing and control, sensing and control technologies for human–robot interaction, and interdisciplinary teaching.
\end{IEEEbiography}

\end{document}